\documentclass[twocolumn]{article}

\usepackage[top=0cm, bottom=1cm, left = 1.5cm, right = 0.5cm,columnsep=20pt]{geometry}
\usepackage{mathpazo}           % Use mathpazo font
\usepackage[T1]{fontenc}
\usepackage[protrusion=true]{microtype} % For nicer spacing
\usepackage[USenglish]{babel}       % For proper line-splits
\usepackage[square,numbers]{natbib} % For references
 
\usepackage{mathtools}      % For \eqref
\usepackage{mathrsfs}       % To number only referenced eqns
\mathtoolsset{showonlyrefs} % To number only referenced eqns

\usepackage{hyperref}       % For reference links

\usepackage{xcolor}         % For colored reference links
\definecolor{zaffre}{cmyk}{1,0.88,0,.34}
\hypersetup{colorlinks, linkcolor={zaffre},citecolor={zaffre},urlcolor={zaffre}}

\usepackage{etoolbox}       % To patch \quote to change margins (next line)
\patchcmd{\quote}{\rightmargin}{\leftmargin 1em \rightmargin}{}{}

\newcommand{\sso}{\rho}     % Start-State Objective
\newcommand{\co}{f}         % Classical Objective

\title{Classical Policy Gradient: Preserving Bellman's Principle of Optimality}

\author{Philip S. Thomas, Scott M. Jordan, Yash Chandak, Chris Nota, and James Kostas\\\normalsize University of Massachusetts Amherst, College of Information and Computer Sciences}

\date{}

\begin{document}
\maketitle
\thispagestyle{empty} % Remove page number

\noindent In 1954, Richard Bellman wrote \citep{Bellman1954}:
\begin{quote}
    \textit{Principle of Optimality: An optimal policy has the property that whatever the initial state and initial decisions are, the remaining decisions must constitute an optimal policy with regard to the state resulting from the first decisions.}
\end{quote}
This principle of optimality has endured at the foundation of reinforcement learning research, and is central to what remains the classical definition of an optimal policy \citep{SuttonBarto2}. Classical reinforcement learning algorithms like $Q$-learning \citep{Watkins1989} embody this principle by striving to act optimally in every state that occurs, regardless of \textit{when} the state occurs. 

The start-state objective function, $\sso(\theta)=\mathbf{E}[\sum_{t=0}^\infty \gamma^t R_t|\theta]$, prioritizes making decisions optimally \emph{in the initial state}, not necessarily in the states resulting from the first decisions.\footnote{We adopt notational standard MDPNv1 \citep{Thomas2016}.} These two goals (optimizing decisions in the initial state and optimizing decisions in subsequent states) can be conflicting when using function approximation, particularly when $\gamma$ is small and the initial state distribution has limited support. So, maximizing $\sso$ does not preserve the principle of optimality. 

Let $q_\theta(s,a)=\mathbf{E}[\sum_{k=0}^\infty \gamma^k R_{t+k} | S_t{=}s,A_t{=}a,\theta]$ so that 
\begin{equation}
    \label{eq:pg}
    \nabla \sso(\theta)=\mathbf{E}\left [ \sum_{t=0}^\infty \gamma^t q_\theta(S_t,A_t) \frac{\partial \ln\big (\pi(S_t,A_t,\theta)\big )}{\partial \theta}\right ].
\end{equation}
The $\gamma^t$ term in \eqref{eq:pg} discounts the importance of optimal behavior in states that occur at later times. Algorithms purported to update $\theta$ following estimates of $\nabla \sso(\theta)$ typically drop this $\gamma^t$ term, since including it or setting $\gamma=1$ results in poor performance. As a result, these algorithms do not capture the essence of $\sso$, do not maximize $\sso$, and are not stochastic gradient algorithms \citep{Nota2019}. 

We propose a different objective function for finite-horizon episodic Markov decision processes that better captures the principal of optimality, and provide an expression for its gradient. This new objective, which we call the \emph{classical objective function}, has the form $\co(\theta)=\sum_{s\in\mathcal S} d_\theta(s) v_\theta(s)$, where $d_\theta$ is a distribution over $\mathcal S$ and $v_\theta(s)=\mathbf{E}[\sum_{k=0}^\infty \gamma^k R_{t+k}|S_t{=}s,\theta]$. This form harks back to the classical definition of an optimal policy, particularly if $d_\theta$ has full support on $\mathcal S$ and does not depend on $\theta$, in which case $\co$ preserves the partial ordering on policies used in the classical definition of an optimal policy.

In model-free reinforcement learning, the agent is not free to sample states from an arbitrary distribution, which makes estimating $\co$ or its gradient challenging with such a $d_\theta$. So, we trade-off similarity to the classical definition of an optimal policy with the practicality of estimating the objective function and its gradient, and define $d_\theta$ to be the \emph{on-policy distribution for episodic tasks} \citep[page 199]{SuttonBarto2}, but with some probability shifted to the terminal absorbing state: $d_\theta(s)=\frac{1}{h}\sum_{t=0}^{h-1} \Pr(S_t{=}s|\theta)$, where $h$ is the horizon. This captures the spirit of classical algorithms like $Q$-learning using function approximation: updates to function approximators occur when states are encountered, and are not discounted.

In the supplementary material we show that 
\begin{equation}
    \nabla \co(\theta)=\mathbf{E} \left [ \frac{1}{h} \sum_{t=0}^{h-1}
    q_\theta(S_t,A_t)
    \sum_{i=0}^{t} w(i,t)  \frac{\partial \ln\big(\pi(S_i,A_i,\theta)\big)}{\partial \theta}\middle | \theta\right ],
\end{equation}
where $w(i,t)=1$ if $i \neq t$, $w(i,t)=(1-\gamma^{t+1})/(1-\gamma)$ if $i=t$ and $\gamma < 1$, and $w(i,t)=t+1$ if $i=t$ and $\gamma = 1$.

The techniques that make estimation of $\nabla \sso$ effective, and which have been developed over 27 years \citep{Williams1992}, do not necessarily carry over to estimating $\nabla \co$. For example, it is not clear how baselines and control variates (and thus actor-critics) should be leveraged. Developing practical algorithms for (approximately) maximizing $\co$ is an open problem---we have only had success with simple REINFORCE-like algorithms.

Notice that $\co$ is not an ideal objective since, like $\sso$, it does not preserve the partial ordering on policies used in the classical definition of an optimal policy, and examples exist wherein it prescribes unreasonable behavior. Still, $\co$ presents a new direction for policy gradient research, opening new questions like: \textbf{1)} are policy gradient algorithms for $\sso$ that drop the $\gamma^t$ term better viewed as algorithms for optimizing $\co$? \textbf{2)} How should baselines and control variates be leveraged when optimizing $\co$? \textbf{3)} Can practical (linear-time and generalized \citep{Thomas2014b}) natural gradient algorithms be derived?\footnote{Our experiments with such methods have hitherto been unsuccessful.} \textbf{4)} Do alternate forms for $\nabla \co$ facilitate gradient estimation, e.g., writing the $t$-summation over $\partial \ln(\pi(S_t,A_t,\theta))/\partial \theta$ and the inner $i$-summation over $q_\theta(S_i,A_i)$ so that the $i$-summation can be expressed as a new value function that measures the expected sum of state-values rather than the expected sum of rewards---a value function that might be approximated using a new TD-like algorithm, and which might allow for actor-critics for the classical objective? \textbf{5)} What are the relationships between $\co$, $\sso$, and the average reward objective? For example, notice that when $\gamma=0$, $\co$ is equivalent to $\sso$ with $\gamma=1$.

\vspace{0.1cm}
\hrule                              % Delineate transition to references
\renewcommand{\bibsection}{}        % Cut "References" title to save space
\setlength{\bibsep}{0pt plus 0.3ex} % Decrease line spacing
\begin{footnotesize}                % Shrink references

\bibliographystyle{unsrtnat}
\end{footnotesize}

\end{document}

\maketitle
\thispagestyle{empty} % Remove page number

\vspace{-1cm}Here we derive the expression for $\nabla \co(\theta)$ presented in the main document.
\begin{align}
    \nabla \co(\theta) =& \sum_{s \in \mathcal S} \frac{\partial}{\partial \theta} d_\theta(s) v_\theta(s)
    =\underbrace{\sum_{s \in \mathcal S} d_\theta(s) \frac{\partial}{\partial \theta} v_\theta(s)}_{\text(a)} +\underbrace{\sum_{s \in \mathcal S} v_\theta(s) \frac{\partial}{\partial \theta} d_\theta(s)}_\text{(b)}.
    \label{eq:productRule}
\end{align}
We will derive expressions for the two terms in \eqref{eq:productRule} independently and then sum them to obtain an expression for $\nabla \co(\theta)$. 
We begin with term \textbf{(a)} in \eqref{eq:productRule}, and start by using a property derived by \citet{Sutton2000}:
\begin{equation}
\label{eq:lkjahwglkjag}
    \forall s \in \mathcal S,\forall t \in \mathbb N_{\geq 0},\, \frac{\partial}{\partial \theta} v_\theta(s)=\sum_{k=0}^{h-1} \sum_{x \in \mathcal S}\gamma^k \Pr(S_{t+k}{=}x|S_t{=}s,\theta)\sum_{a \in \mathcal A} q_\theta(x,a) \frac{\partial \pi(x,a,\theta)}{\partial \theta},
\end{equation}
which implies that for all $t \in \mathbb N_{\geq 0}$,
\begin{align}
    \sum_{s \in \mathcal S} d_\theta(s) \frac{\partial}{\partial \theta} v_\theta(s)
    =&
    \sum_{s \in \mathcal S}d_\theta(s)  \sum_{k=0}^{h-1}\sum_{x \in \mathcal S} \gamma^k \Pr(S_{t+k}{=}x | S_t{=}s, \theta) \sum_{a \in \mathcal A} q_\theta(x,a) \frac{\partial \pi(x,a,\theta)}{\partial \theta}\\
    =&\sum_{s \in \mathcal S}\frac{1}{h}\sum_{t=0}^{h-1} \Pr(S_t{=}s|\theta)  \sum_{k=0}^{h-1}\sum_{x \in \mathcal S}\sum_{a \in \mathcal A} \gamma^k \Pr(S_{t+k}{=}x | S_t{=}s, \theta)   q_\theta(x,a) \pi(x,a,\theta)\frac{\partial \ln (\pi(x,a,\theta))}{\partial \theta}\\
    =&\frac{1}{h}\sum_{s \in \mathcal S} \sum_{t=0}^{h-1}\sum_{k=0}^{h-1}\sum_{x \in \mathcal S} \sum_{a \in \mathcal A} \Pr(S_t{=}s|\theta)    \Pr(S_{t+k}{=}x | S_t{=}s, \theta) \Pr(A_{t+k}{=}a |S_{t+k}{=}x,\theta) \gamma^k q_\theta(x,a) \frac{\partial \ln(\pi(x,a,\theta))}{\partial \theta}\\
    =&\frac{1}{h} \sum_{t=0}^{h-1}\sum_{k=0}^{h-1}\sum_{x \in \mathcal S} \sum_{a \in \mathcal A} \Pr(S_{t+k}{=}x |  \theta) \Pr(A_{t+k}{=}a |S_{t+k}{=}x,\theta) \gamma^k q_\theta(x,a) \frac{\partial \ln(\pi(x,a,\theta))}{\partial \theta},
\end{align}
since $\Pr(A_{t+k}{=}a |S_{t+k}{=}x,\theta)=\Pr(A_{t+k}{=}a |S_{t+k}{=}x,S_t{=}s,\theta)$ and by the law of total probability. 
Continuing, starting with the fact that $\Pr(S_{t+k}{=}x |  \theta) \Pr(A_{t+k}{=}a |S_{t+k}{=}x,\theta)= \Pr(S_{t+k}{=}x,A_{t+k}{=}a |\theta)$, we have that:
\begin{align}
    \sum_{s \in \mathcal S} d_\theta(s) \frac{\partial}{\partial \theta} v_\theta(s) =&\frac{1}{h}\sum_{t=0}^{h-1} \sum_{k=0}^{h-1} \sum_{x \in \mathcal S} \sum_{a \in \mathcal A}  \Pr(S_{t+k}{=}x, A_{t+k}{=}a | \theta)  \gamma^k  q_\theta(x,a) \frac{\partial \ln(\pi(x,a,\theta))}{\partial \theta}\\
    =&\frac{1}{h}\sum_{t=0}^{h-1} \sum_{k=0}^{h-1}   \mathbf{E} \left [  \gamma^k q_\theta(S_{t+k},A_{t+k}) \frac{\partial \ln(\pi(S_{t+k},A_{t+k},\theta))}{\partial \theta} \middle | \theta \right ]\\
    =&\frac{1}{h}\sum_{t=0}^{h-1} \sum_{i=t}^{h-1+t} \mathbf{E} \left [  \gamma^{i-t} q_\theta(S_{i},A_{i}) \frac{\partial \ln(\pi(S_i,A_i,\theta))}{\partial \theta} \middle | \theta \right ],
    \end{align}
    by substitution of the variable $i=t+k$. 
    Since $S_i$ is the terminal absorbing state for $i \geq h$, we have that $q_\theta(S_i,a_i)=0$ for $i \geq h$, and thus the sum over $i$ can stop at $h-1$ rather than $h-1+t$. 
    Continuing, starting with this change and then reordering the summations over $t$ and $i$, we have:
    \begin{align}
    \sum_{s \in \mathcal S} d_\theta(s) \frac{\partial}{\partial \theta} v_\theta(s) =&\frac{1}{h}\sum_{t=0}^{h-1} \sum_{i=t}^{h-1} \mathbf{E} \left [  \gamma^{i-t} q_\theta(S_{i},A_{i}) \frac{\partial \ln(\pi(S_i,A_i,\theta))}{\partial \theta} \middle | \theta \right ]\\
    =& \mathbf{E} \left [  \frac{1}{h}\sum_{i=0}^{h-1} \sum_{t=0}^{i} \gamma^{i-t} q_\theta(S_{i},A_{i}) \frac{\partial \ln(\pi(S_i,A_i,\theta))}{\partial \theta} \middle | \theta \right ]\\
    =& \mathbf{E} \left [  \frac{1}{h}\sum_{i=0}^{h-1} w(i,i) q_\theta(S_{i},A_{i}) \frac{\partial \ln(\pi(S_i,A_i,\theta))}{\partial \theta}\middle | \theta \right ],
\end{align}
since $\sum_{t=0}^{i} \gamma^{i-t}$ is equal to $w(i,i)$, which is  $\frac{1-\gamma^{i+1}}{1-\gamma}$ if $\gamma < 1$ and $i+1$ otherwise. 
Replacing the symbol $i$ with the symbol $t$ we have:
\begin{align}
\label{eq:term1comb}
    \sum_{s \in \mathcal S} d_\theta(s) \frac{\partial}{\partial \theta} v_\theta(s)
    =& \mathbf{E} \left [  \frac{1}{h}\sum_{t=0}^{h-1} w(t,t) q_\theta(S_t,A_t) \frac{\partial \ln(\pi(S_t,A_t,\theta))}{\partial \theta} \middle | \theta \right ].
\end{align}
Notice that \eqref{eq:term1comb} closely resembles the policy gradient for the start-state setting with the $\gamma^t$ term removed. 
Also notice that the left side of \eqref{eq:term1comb} captures how changes to the policy parameters change the value of states, but does not capture how changes to the policy parameters change the state distribution. 
This suggests that removing the $\gamma^t$ term from the policy gradient theorem for the start-state objective function results in an update that does not properly account for how changes to the policy change the state distribution.

We now simplify term \textbf{(b)} in \eqref{eq:productRule}. 
Let $T_t$ be a trajectory (a random variable) that includes the states and actions (not the rewards) up until (and including) time $t$. 
That is $T_t=(S_0,A_0,S_1,A_1,\dotsc,S_t,A_t)$. 
Let $\mathcal T_t$ be the set of all possible values for $T_t$. 
Using this notation, we simplify term \textbf{(b)} in \eqref{eq:productRule}: 
\begin{align}
    \sum_{s \in \mathcal S} v_\theta(s) \frac{\partial}{\partial \theta} d_\theta(s)
    =&\sum_{s \in \mathcal S} v_\theta(s) \frac{\partial}{\partial \theta} \frac{1}{h} \sum_{t=0}^{h-1} \Pr(S_t{=}s|\theta)\\
    =& \frac{1}{h} \sum_{s \in \mathcal S} v_\theta(s)  \frac{\partial}{\partial \theta}\sum_{t=0}^{h-1} \sum_{\tau_{t-1} \in \mathcal T_{t-1}}\!\!\!\!\Pr(T_{t-1}{=}\tau_{t-1},S_t{=}s|\theta)\\ 
    =& \frac{1}{h}\sum_{t=0}^{h-1} \sum_{s_t \in \mathcal S}  \sum_{\tau_{t-1} \in \mathcal T_{t-1}} \!\!\!\! v_\theta(s_t)  \frac{\partial}{\partial \theta}\Pr(T_{t-1}{=}\tau_{t-1},S_t{=}s_t|\theta). 
\end{align}
Continuing, with $\tau_{t-1}=(s_0,a_0,s_1,a_1,\dotsc,s_{t-1},a_{t-1})$, using $p$ to denote the state-transition function, and writing $p(s_{-1},a_{-1},s_0)$ to denote $\Pr(S_0{=}s_0)$, we have:
\begin{align}
    \sum_{s \in \mathcal S} v_\theta(s) \frac{\partial}{\partial \theta} d_\theta(s)
    =& \frac{1}{h} \sum_{t=0}^{h-1} \sum_{s_t\in \mathcal S}\sum_{\tau_{t-1} \in \mathcal T_{t-1}}v_\theta(s_t)\frac{\partial}{\partial \theta} \left [ \left ( \prod_{i=0}^{t-1} p(s_{i-1}, a_{i-1}, s_i) \pi(s_i,a_i,\theta)\right ) p(s_{t-1},a_{t-1},s_t) \right ]\\
    =& \frac{1}{h} \sum_{t=0}^{h-1} \sum_{s_t\in \mathcal S}\sum_{\tau_{t-1} \in \mathcal T_{t-1}}v_\theta(s_t)  \left (\prod_{i=0}^{t} p(s_{i-1}, a_{i-1}, s_i) \right )
    \frac{\partial}{\partial \theta}  \prod_{i=0}^{t-1}  \pi(s_i,a_i,\theta)\\
    =& \frac{1}{h} \sum_{t=0}^{h-1} \sum_{s_t\in \mathcal S}\sum_{\tau_{t-1} \in \mathcal T_{t-1}}  \left (\prod_{i=0}^{t} p(s_{i-1}, a_{i-1}, s_i) \right )
    \left (\prod_{i=0}^{t-1}  \pi(s_i,a_i,\theta)\right ) v_\theta(s_t) \sum_{i=0}^{t-1} \frac{\partial}{\partial \theta} \ln(\pi(s_i,a_i,\theta))\\
    =& \frac{1}{h} \sum_{t=0}^{h-1} \sum_{s_t\in \mathcal S}\sum_{\tau_{t-1} \in \mathcal T_{t-1}}  \Pr(T_{t-1}{=}\tau_{t-1},S_t{=}s_t|\theta)v_\theta(s_t) \sum_{i=0}^{t-1}\frac{\partial \ln(\pi(S_i,A_i,\theta))}{\partial \theta}\\
    =& \frac{1}{h} \sum_{t=0}^{h-1} \mathbf{E}\left [v_\theta(S_t)   \sum_{i=0}^{t-1} \frac{\partial \ln(\pi(S_i,A_i,\theta))}{\partial \theta} \middle | \theta \right ]\\
    =& \mathbf{E}\left [\frac{1}{h} \sum_{t=0}^{h-1}  v_\theta(S_t)   \sum_{i=0}^{t-1} \frac{\partial \ln(\pi(S_i,A_i,\theta))}{\partial \theta} \middle | \theta \right ]\\
    =& \mathbf{E}\left [\frac{1}{h} \sum_{t=0}^{h-1}  q_\theta(S_t,A_t)   \sum_{i=0}^{t-1} \frac{\partial \ln(\pi(S_i,A_i,\theta))}{\partial \theta} \middle | \theta \right ].
    \label{eq:term2comb}
\end{align}
Summing \eqref{eq:term1comb} and \eqref{eq:term2comb} and using the fact that $w(i,t)=1$ if $i \neq t$, we obtain an expression for the sum of terms \textbf{(a)} and \textbf{(b)} in \eqref{eq:productRule}: 
\begin{align}
    \nabla \co(\theta)=& \mathbf{E} \left [  \frac{1}{h}\sum_{t=0}^{h-1} w(t,t) q_\theta(S_t,A_t) \frac{\partial \ln(\pi(S_t,A_t,\theta))}{\partial \theta} \middle | \theta \right ] + \mathbf{E}\left [\frac{1}{h} \sum_{t=0}^{h-1}  q_\theta(S_t,A_t)   \sum_{i=0}^{t-1} \frac{\partial \ln(\pi(S_i,A_i,\theta))}{\partial \theta} \middle | \theta \right ]\\
    =&\mathbf{E} \left [  \frac{1}{h}\sum_{t=0}^{h-1} q_\theta(S_t,A_t) w(t,t)  \frac{\partial \ln(\pi(S_t,A_t,\theta))}{\partial \theta} + \frac{1}{h} \sum_{t=0}^{h-1}  q_\theta(S_t,A_t)   \sum_{i=0}^{t-1} w(i,t) \frac{\partial \ln(\pi(S_i,A_i,\theta))}{\partial \theta}\middle | \theta \right ]\\
    =&\mathbf{E} \left [  \frac{1}{h}\sum_{t=0}^{h-1} q_\theta(S_t,A_t) \sum_{i=0}^t w(i,t)  \frac{\partial \ln(\pi(S_i,A_i,\theta))}{\partial \theta}\middle | \theta \right ].
\end{align}
\thispagestyle{empty}               % Remove page number

\bibliographystyle{abbrvnat}